# Collaborative Discrepancy Optimization for Reliable Image Anomaly Localization

Yunkang Cao, *Student Member, IEEE,* Xiaohao Xu, Zhaoge Liu, Weiming Shen, *Fellow, IEEE*

*Abstract*—Most unsupervised image anomaly localization methods suffer from overgeneralization because of the high generalization abilities of convolutional neural networks, leading to unreliable predictions. To mitigate the overgeneralization, this study proposes to collaboratively optimize normal and abnormal feature distributions with the assistance of synthetic anomalies, namely collaborative discrepancy optimization (CDO). CDO introduces a margin optimization module and an overlap optimization module to optimize the two key factors determining the localization performance, *i.e.*, the margin and the overlap between the discrepancy distributions (DDs) of normal and abnormal samples. With CDO, a large margin and a small overlap between normal and abnormal DDs are obtained, and the prediction reliability is boosted. Experiments on MVTec2D and MVTec3D show that CDO effectively mitigates the overgeneralization and achieves great anomaly localization performance with real-time computation efficiency. A real-world automotive plastic parts inspection application further demonstrates the capability of the proposed CDO. Code is available on https://github.com/caoyunkang/CDO.

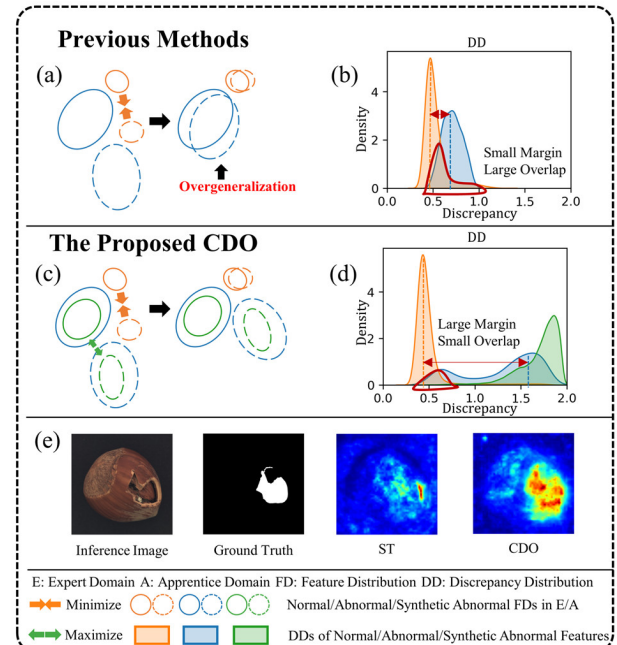

Fig. 1. Comparison between previous methods and the proposed CDO. Previous methods only minimize the discrepancies between normal FDs in the expert and apprentice domains, bringing small discrepancies for some abnormal samples and leading to high prediction uncertainty, thus the overgeneralization problem. Instead, the proposed CDO collaboratively optimizes the normal and abnormal discrepancy distributions with the assistance of synthesized anomalies to mitigate the overgeneralization problem.

## I. INTRODUCTION

Image anomaly localization aims to localize anomalies or defects that significantly deviate from normal patterns. Its abnormalities localizing ability can potentially greatly improve the effectiveness and performance of product quality controls [1]–[4] and thermal image-based fault diagnosis [5], which makes itself crucial in intelligent manufacturing systems.

Generally, anomaly samples are hard to collect in real-world applications, especially in the early deployment stage. It is also costly to make precise pixel-level annotations for anomaly samples. The lack of precisely annotated samples poses a challenge to the application of broadly investigated supervised semantic segmentation methods in the anomaly localization task. To mitigate the requirement for large amounts of annotated data, unsupervised image anomaly localization methods that only require normal samples in the training phase have gained significant popularity.

Most unsupervised image anomaly localization methods [6]–[11] localize anomalies in the same fashion. In the training stage, they try to train a model to describe the normal feature distribution (FD). In the inference stage, they localize abnormal features according to the response of the model to the testing features. Among existing unsupervised image anomaly localization methods, reconstruction-based [12]–[17] and knowledge distillation-based [18]–[23] methods describe the normal FD implicitly with a feature-to-feature discrepancy optimization scheme. This scheme contains an informative expert and an initially-awkward apprentice, and both can extract features for input data and construct respective domains. In the expert domain, features distribute discriminatively, and normal and abnormal features can be distinguished well. To describe the normal FD, the apprentice is trained to mimic the normal FD in the expert domain by minimizing the feature-to-feature discrepancies between the two normal FDs, as shown in Fig. 1 (a). In existing methods [12]–[21], it is commonly assumed that the discrepancy minimization between the normal FDs rarely affects the abnormal FDs, and abnormal FDs still distribute differently. Hence, the discrepancies between features corresponding to the same input data in the two domains indicate the anomaly degree.

However, the assumption is only partially valid because of the high generalization ability of the apprentice. When minimizing discrepancies between normal FDs, the apprentice may generalize unexpectedly and produce a similar abnormal FD to that in the expert domain, as Fig. 1 (a) shows, resulting in some low discrepancies between the abnormal FDs. This unexpected generalization is defined as the overgeneralization problem in this paper. The overgeneralization problem seriously affects the discrepancy distribution (DD) between corresponding features in the expert and apprentice domains, and the margin and the overlap between DDs of normal and abnormal features are also damaged. The margin and the overlap between DDs of normal features and abnormal features are two key factors that influence anomaly localization performance. The margin can be roughly defined as the absolute difference between the average values of the two DDs. The overlap can be regarded as the percentage of parts having the same values. The overgeneralization problem causes an overall decrease in the discrepancies between abnormal FDs, which results in a small margin and a large overlap between the DDs of normal features and abnormal features and makes the anomaly localization predictions less reliable, as Fig. 1 (b) shows. Whereas existing unsupervised image anomaly localization methods have achieved comparable localization performance, they only minimize the discrepancies between normal FDs and rarely address the overgeneralization, which leaves considerable further improvements for image anomaly localization methods.

To address the overgeneralization of existing unsupervised anomaly localization methods and ensure their prediction reliability, this study proposes a method to optimize the discrepancies between the normal FDs and abnormal FDs collaboratively, namely *Collaborative Discrepancy Optimization (CDO)*, as Fig. 1 (c) shows. With CDO, the apprentice is aware of not only mimicking the normal FD in the expert domain but also producing an easily-distinguished abnormal FD to that in the expert domain, thus mitigating the overgeneralization problem. Concretely, CDO alleviates the overgeneralization problem by minimizing the discrepancies between the normal FDs and maximizing the abnormal FDs in the expert and apprentice domains simultaneously. The simultaneous optimization avoids those low discrepancies between abnormal FDs and improves the prediction reliability. However, CDO cannot directly maximize the discrepancies between abnormal FDs due to the inaccessibility of abnormal data in the training phase. Hence, CDO proposes to generate synthetic abnormal inputs via random perturbation and maximize the discrepancies between the two synthetic abnormal FDs instead of the real abnormal FDs. The maximization implicitly pushes the real abnormal FDs away, and assists in acquiring large discrepancies between abnormal FDs, as Fig. 1 (c) shows, which empowers the apprentice with a stronger ability to localize anomalies.

Specifically, CDO proposes to optimize the two critical factors, the margin and the overlap between the DDs, respectively, and consists of a margin optimization module (MOM) and an overlap optimization module (OOM). MOM minimizes the discrepancies of normal FDs and maximizes those of synthetic abnormal FDs simultaneously to optimize the margin. To optimize the overlap, OOM raises attention to the tailed hard samples that are the most significant ones influencing the overlap, namely normal samples with large discrepancies and abnormal ones with small discrepancies. OOM assigns weights to individual samples dynamically based on the statistical information of DDs. Those tailed samples are assigned larger weights to draw more attention in the training process for the overlap reduction. With the proposed MOM and OOM, a large margin and a small overlap between the DDs of normal features and abnormal features are obtained, achieving reliable anomaly localization performance, as Fig. 1 (d) shows. Fig. 1 (e) also shows that CDO achieves significantly better results than its strong baseline ST [20].

The contributions of this study can be summarized as follows:
● This study proposes a general formulation that takes anomaly localization into a discrepancy optimization scheme. Based on this formulation, this study presents Collaborative Discrepancy Optimization (CDO), which produces reliable and excellent anomaly localization.
● This study proposes a margin optimization module (MOM) and an overlap optimization module (OOM) to support the proposed CDO, obtaining a large margin and a small overlap between the normal and abnormal discrepancy distributions.
● For further validation, this study collects a real-world automotive plastic part dataset with pixel-wise annotations. Compared to existing anomaly detection datasets, the collected dataset is more challenging because it has normal patches with larger inter-class variance and anomalies of extremely small size. We applied the proposed CDO to this dataset and achieved impressive anomaly localization performance.

The remainder of the paper is organized as follows. Section II comprehensively reviews the related work on unsupervised image anomaly localization. CDO is illustrated in detail in Section III. Section IV presents thorough experiments, ablation studies, results, and analyses. Finally, Section V concludes this paper and discusses future research directions.

## II. RELATED WORK

Unsupervised image anomaly localization methods can be categorized into three types: distribution-based [6]–[10], reconstruction-based [12]–[17], and knowledge distillation-based methods [18]–[21].

### A. Distribution-based methods

Distribution-based methods [6]–[10] directly leverage clustering models or generative models to describe the normal FD. For example, SPADE [10] built a feature gallery first and then dynamically retrieved the most similar features in the gallery to the tested one using KNN. The distances between those features are used to score anomalies. However, SPADE had an inefficient computation as two KNNs were needed. Instead, GCPF [8] used several Gaussian clusters to describe the normal FD, and the computation speed was significantly improved. Similarly, PaDiM [9] leveraged individual Gaussian distribution in every location and achieved better performance, but PaDiM still required large memory consumption and performed subpar when images were not aligned well. DifferNet [6] and CFLOW [7] were based on a normalizing flow framework, which can automatically describe the distribution of normal features and explicitly estimate the likelihood of the tested features. Despite some improvements in anomaly localization, they are often time-consuming since the pre-trained network and decoder typically work sequentially.

*B. Reconstruction-based methods*

Reconstruction-based methods [12]–[17] attempt to reconstruct the corresponding normal features for arbitrary noisy input features. The reconstructed features act as the apprentice domain, mimicking the expert domain constructed by input features. AESSIM [17] reconstructed the normal RGB values under several similarity constraints. MemAE [16] reconstructed normal features from a gallery, offering stronger constraints to prevent the reconstruction of abnormal features. Since RGB values are not distinctive enough, DFR [13] used pre-trained high-dimension features as inputs and achieved better performance.

Moreover, DRAEM [14], RIAD [12], and dual-Siamese network [15] also generated anomalies to achieve better localization performance, but their purpose is virtually different from the proposed CDO. They generated anomalies to explicitly endow the network with the capability to suppress anomalous for better reconstruction quality. Thus, they try to output normal features under arbitrary inputs. In other words, they try to minimize the discrepancies between the normal FD in the expert domain and the normal and abnormal FDs in the apprentice domain, which is quite difficult and still cannot guarantee low prediction uncertainty. By contrast, CDO generates anomalies to optimize the margin and overlap between the two DDs, effectively decreasing the prediction uncertainty and improving anomaly localization performance.

*C. Knowledge distillation-based methods*

Knowledge distillation-based methods [18]–[23] provide a concise scheme to implicitly model the normal FD. In this scheme, the features from the pre-trained teacher network compose the expert domain, and those from the student network construct the apprentice domain. During the training phase, the student network only learns to regress the normal features, so it would deviate under abnormal features, and the pixel-wise feature regression errors indicate the anomaly scores. US [19] and MRKD [18] were the first to detect anomalies based on knowledge distillation. US [19] investigated multi-resolution feature representation by cropping patches with different sizes. Instead, MRKD [18] leveraged the features from different hierarchies to capture the multi-resolution context. ST [20] developed further with feature pyramid matching and achieved effective yet powerful anomaly localization performance. IKD [21] found high overfitting risk existed in the knowledge distillation-based scheme because of the mismatch between network capacity and knowledge contained in homogeneous images. IKD [21] proposed to distill informative knowledge and alleviate the overfitting. RD4AD [22] noticed the overgeneralization problem and mitigated it by reverse distillation, which is less likely to be overgeneralized as the expert and apprentice networks have different architectures. SKD [23] concluded that some real anomaly samples contribute to better localization performance. The aforementioned methods have reached state-of-the-art localization performance with great efficiency. However, knowledge distillation-based methods only minimize the discrepancies between the normal FDs. Due to the overgeneralization problem, the discrepancies between abnormal FDs also decrease during the discrepancy minimization of normal FDs, leading to high prediction uncertainties.

This study proposes to optimize the normal and abnormal FDs collaboratively, and the proposed CDO enlarges the margin and decreases the overlap between the normal and abnormal DDs, producing reliable and excellent anomaly localization performance.

### III. COLLABORATIVE DISCREPANCY OPTIMIZATION

*A. Problem Definition*

The image anomaly localization task aims at localizing abnormal regions that deviate from normal patterns and can be formulated as follows. $\mathbb{P} = \{p_1, p_2, ..., p_N\}$ is a set of pixels of anomaly-free training images. Image anomaly localization methods localize abnormal regions by learning an anomaly scoring function: $g: p \to \mathbb{R}$ that assigns large anomaly scores to abnormal pixels and relatively low scores to normal pixels.

In the rest of this section, this paper first describes a general formulation of existing methods. Then, the details of the proposed CDO are elaborated. The process of anomaly score calculation is depicted in the end.

*B. General Formulation of Existing Methods*

Denoting the normal and abnormal FDs corresponding to respective pixels in the expert and the apprentice domains as $E(p_n)$, $E(p_a)$, $A(p_n)$, and $A(p_a)$, most existing unsupervised anomaly localization methods can be formulated as the process of minimizing the discrepancies between $E(p_n)$ and $A(p_n)$, which can be formulated as follows:

$$J = D\big(E(p_n), A(p_n)\big) \quad (1)$$

where $D(\cdot,\cdot)$ denotes a function for feature-to-feature discrepancy calculation between two distributions. Previous work [12]–[21] assumed that the discrepancies between $E(p_a)$ and $A(p_a)$ would not be significantly affected and keep large during the minimization of discrepancies between $E(p_n)$ and $A(p_n)$. However, because of the overgeneralization, the discrepancies between $E(p_a)$ and $A(p_a)$ may also decrease, resulting in high prediction uncertainty.

To tackle the overgeneralization problem, this study proposes CDO, which improves the prediction certainty by optimizing discrepancies between normal FDs and abnormal FDs collaboratively. As no abnormal pixels can be accessed in the training stage, CDO introduces to leverage synthetic abnormal pixels for the assistance of discrepancies maximization between $E(p_a)$ and $A(p_a)$. As Fig. 2 shows, CDO contains three modules: discrepancy distribution generation (DDG), margin optimization module (MOM), and overlap optimization module (OOM). Firstly, CDO perturbs the input normal image and maps the perturbed image into the expert and apprentice domains with respective map functions. Normal and synthetic abnormal features in the two domains are also acquired. Secondly, with the discrepancy function, DDs of normal features and synthetic abnormal features are generated. Thirdly, OOM assigns weights to individual features according to their discrepancies to raise more attention to hard tailed samples. Finally, to enlarge the margin between DDs, MOM maximizes the discrepancies of synthetic abnormal FDs and minimizes those of normal FDs collaboratively under the weights produced by the OOM. DDG, MOM, and OOM will be described in the following subsections.

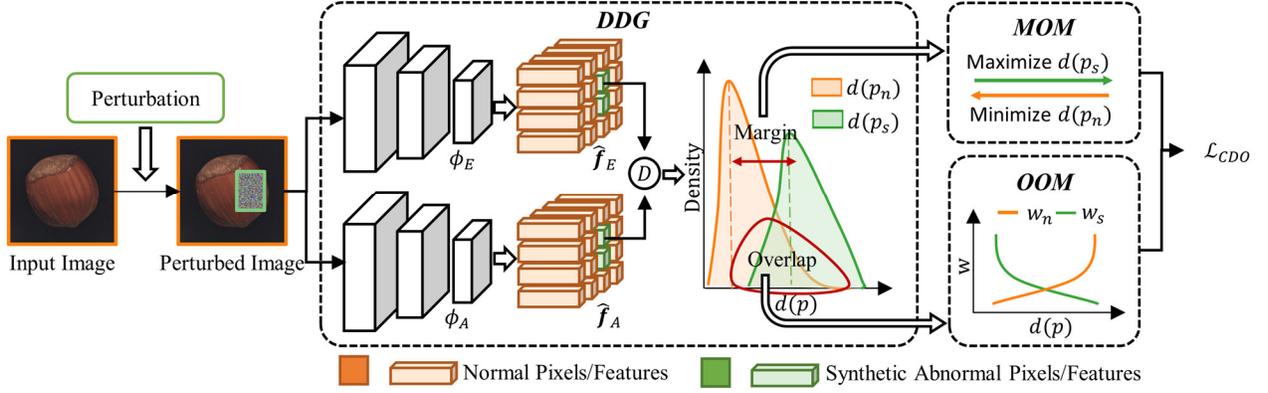

Fig. 2. The framework of CDO.

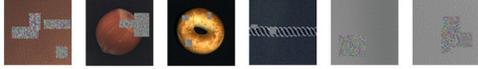

Fig. 3. Some samples of perturbed images.

### C. DDG

The normal and abnormal discrepancies between the two domains are formulated as $d(p_n)$ and $d(p_a)$:
$$d(p) = D(\phi_E(p;\theta_E), \phi_A(p;\theta_A)) \quad (2)$$
where $p_n$ and $p_a$ denote normal and abnormal pixels; $\phi_E$ and $\phi_A$ denote the map functions that can map pixels into respective domains and extract features; $\theta_E$ and $\theta_A$ are the corresponding parameters for the expert and apprentice networks, and $\theta_E$ keeps frozen while $\theta_A$ is optimized during the training phase.

This study implements CDO based on the knowledge distillation-based scheme [18]–[21] because this scheme has achieved excellent performance and high computation efficiency. Concretely, in our implementation, $\phi_E$ denotes a pre-trained network, and $\phi_A$ denotes the corresponding randomly initialized network. The features for respective pixels extracted by $\phi_E$ and $\phi_A$ are denoted as $f_E$ and $f_A$; $\hat{f}_E$ and $\hat{f}_A$ denote the normalized features. Pixel-wise mean square errors are used to evaluate the discrepancy between the two features,
$$d(p) = D(\hat{f}_E, \hat{f}_A) = \|\hat{f}_E - \hat{f}_A\|_2 \quad (3)$$

This study proposes to synthesize anomalies via random perturbation and maximize the discrepancies between them to optimize the discrepancies between features of real anomalies implicitly. Specifically, several squares are randomly generated, and the corresponding regions are replaced with random values sampled from a Gaussian normal distribution. Some samples of perturbed images are shown in Fig. 3. With random perturbation, normal and synthetic abnormal pixel sets $\{p_n\}$ and $\{p_s\}$ can be collected, and the numbers of pixels are denoted as $N_n$ and $N_s$, respectively. The generated DDs are denoted as $\{d(p_n)\}$ and $\{d(p_s)\}$.

### D. MOM

Existing methods only minimized $\{d(p_n)\}$ and neglected the optimization of $\{d(p_s)\}$,
$$\mathcal{L} = \frac{\sum_{i=0}^{N_n} d(p_n)_i}{N_n} \quad (4)$$

Despite their impressive performances, high prediction uncertainty exists because of the overgeneralization problem. To address the overgeneralization problem, MOM minimizes $\{d(p_n)\}$ and maximizes $\{d(p_s)\}$ simultaneously to enlarge the margin between the two DDs,
$$\mathcal{L}_{MOM} = \frac{\sum_{i=0}^{N_n} d(p_n)_i - \sum_{j=0}^{N_s} d(p_s)_j}{N_n + N_s} \quad (5)$$

The improved optimization objective can empower the apprentice to mimic the expert as well as to be aware of producing large discrepancies when the input pixels are out-of-distribution. With MOM, the margin between the normal and abnormal DDs can be enlarged significantly, and better performance can be produced. However, simply optimizing the average of normal and abnormal discrepancies may not optimize some tail samples well, still resulting in a large overlap between DDs. Hence, OOM is developed to raise specific attention to those tailed samples and complement the drawbacks of solely using MOM.

### E. OOM

With MOM, the DDs of normal features and abnormal features will be endowed with a larger margin than existing methods. While increasing the margin can effectively improve the performance, tailed samples are not optimized well and may affect the overlap. To better reduce the overlap, inspired by the Focal Loss [24], this study introduces OOM to dynamically assign weights to individual pixels during the training stage. OOM focuses on optimizing tailed samples to achieve better overall optimization results. Technically, the ratio between individual discrepancies to the average is a great indicator of pixel-wise importance in the training stage. This study first calculates the averages of normal and abnormal discrepancies, respectively,

**Algorithm 1:** Framework of CDO
**# Training Stage**
**Input:** Training normal pixels $\mathbb{P} = \{p_1, p_2, \ldots, p_N\}$; Expert network and its parameters $\phi_E$, $\theta_E$; Apprentice network and its parameters $\phi_A$, $\theta_A$;
**Output:** Anomaly scoring function $g: p \to \mathbb{R}$
1: Initialize $\theta_E$ with pre-trained weights
2: Initialize $\theta_A$ with random weights
3: **for** $epoch = 1$ to $n\_epochs$ **do**
4:   **for** $batch = 1$ to $n\_batches$ **do**
5:     $\{p\} \leftarrow$ Randomly sample several pixels from $\mathbb{P}$
6:     $\{p_n\}, \{p_s\} \leftarrow$ Randomly perturb $\{p\}$ to generate synthetic abnormal pixels
7:     Compute $\{d(p_n)\}, \{d(p_s)\}$ according to Equation (3)
8:     Compute $\{w_n\}, \{w_s\}$ according to Equation (6) and (7)
9:     Compute $\mathcal{L}_{CDO}$ according to Equation (8)
10:    Perform a gradient descent step to update $\theta_A$
11:   **end for**
12: **end for**
12: Compute $g$ according to Equation (2), (9), and (10)
14: **return** $g$
**# Inference Stage**
**Input:** Unknown pixels $\mathbb{P}_u$, Anomaly scoring function $g: p \to \mathbb{R}$
**Output:** Anomaly scores for $\mathbb{P}_u$
1: Calculate anomaly scores using $g$
2: **return** anomaly scores for $\mathbb{P}_u$

$$\mu_n = \frac{\sum_{i=0}^{N_n} d(p_n)_i}{N_n}, \mu_s = \frac{\sum_{j=0}^{N_s} d(p_s)_j}{N_s} \quad (6)$$

As normal FDs should have significantly smaller discrepancies than those of abnormal FDs, for discrepancies between normal FDs, the larger the ratio to the average, the more attention should be paid. For the discrepancies between abnormal FDs, the smaller the ratio, the more attention should be paid. Therefore, the weights for discrepancies are designed as follows:

$$(w_n)_i = \left[\frac{d(p_n)_i}{\mu_n}\right]^\gamma, (w_s)_i = \left[\frac{d(p_s)_i}{\mu_s}\right]^{-\gamma} \quad (7)$$

where $\gamma \geq 0$ is a hyper-parameter that adjusts the effect of the modulation. Notably, the exponent of normal samples is $\gamma$, and that of abnormal samples is $-\gamma$. Therefore, normal pixels with discrepancies larger than the average are enhanced, and the weights of abnormal pixels with discrepancies smaller than the average are increased. Then the loss function of CDO combined with OOM and MOM can be formulated as:

$$\mathcal{L}_{CDO} = \frac{\sum_{i=0}^{N_n}(w_n)_i d(p_n)_i - \sum_{j=0}^{N_s}(w_s)_j d(p_s)_j}{\sum_{i=0}^{N_n}(w_n)_i + \sum_{j=0}^{N_s}(w_s)_j} \quad (8)$$

### F. Anomaly Score Calculation

After optimizing $\phi_A$ with the proposed CDO, the expert domain $E(p)$ and the apprentice domain $A(p)$ are endowed with small discrepancies between normal FDs and large discrepancies between abnormal FDs. Then the anomaly scoring function $g$ for individual pixels can be defined as:

$$g(p) = d(p) = \left\|\hat{\boldsymbol{f}}_E - \hat{\boldsymbol{f}}_A\right\|_2 \quad (9)$$

where $\hat{\boldsymbol{f}}_E$ and $\hat{\boldsymbol{f}}_A$ denote the normalized features extracted by the expert network $\phi_E$ and apprentice network $\phi_A$.

Moreover, it has been proved that multi-hierarchical representations extracted from hidden layers of convolutional neural networks (CNNs) can further improve anomaly localization performance [8], [9], [20]. Hence, $E(p)$ and $A(p)$ in different hierarchies are leveraged, and the anomaly scores can be further defined as:

$$g(p) = \sum_{i=0}^{\mathcal{H}} d(p)^{(l)} \quad (10)$$

where $d(p)^{(l)}$ denotes the discrepancy between the $l^{\text{th}}$ hierarchy $E(x)$ and $A(x)$, and $\mathcal{H}$ represents the number of hierarchies. Algorithm 1 summarizes the framework of CDO.

## IV. EXPERIMENTS

In this section, several sets of experiments are conducted on MVTec2D [3] and MVTec3D [1] to evaluate the performance of CDO and illustrate the influence of individual components. Besides, CDO is applied to inspect automotive plastic parts to validate its effectiveness in real-world applications.

### A. Experiments Settings

*Dataset Descriptions:* MVTec2D [3] was released by MVTec Software and has been widely studied for unsupervised image anomaly localization. More recently, a new dataset MVTec3D [1] was published, creating a new task to localize mainly geometric anomalies in 3D objects. RGB images and 3D scans are provided in MVTec3D. MVTec2D contains 15 categories, of which five are texture and ten are object categories. MVTec3D contains ten categories, and all of them are object categories. Both MVTec2D and MVTec3D have various anomaly types, such as scratch, colored, and broken. When localizing geometry anomalies in MVTec3D, CDO leverages no 3D scan because CDO only requires RGB images as inputs.

*Evaluation Metrics:* The commonly used area under the receiver operating characteristic curve (AU-ROC) [3] and the normalized area under the per-region overlap curve (AU-PRO) [19] with a threshold of 0.3 is calculated in this study. For both AU-ROC and AU-PRO, a higher value indicates better detection performance.

*Implementation Details:* All images are normalized using the mean and variance of the ImageNet dataset and resized to $256 \times 256$. $\phi_E$ is a network pre-trained on the ImageNet [25] and is frozen during training. $\phi_A$ is a randomly initialized network and is optimized by the AdamW algorithm [26] with a learning rate of 0.0002 and a batch size of 8. Unless otherwise specified, HRNet32 [27] is used as the backbones of $\phi_E$ and $\phi_A$ by default. The coefficient $\gamma$ in Eq. (7) is fixed to 2 by default. Moreover, the expert and apprentice domains from the first three hierarchies are leveraged. The number of training epochs is set as 50. All results reported below are taken from the average and variance of the last five epochs.

TABLE I. QUANTITATIVE COMPARISON OF ANOMALY DETECTION METHODS ON THE MVTEC2D DATASET IN TERMS OF AU-ROC. CDO* AND CFLOW* DENOTE THE EVALUATED RESULTS UNDER THE EVALUATION SETTING OF CFLOW.

| Category | Carpet | Grid | Leather | Tile | Wood | Bottle | Cable | Capsule | Hazelnut | Metal Nut | Pill | Screw | Toothbrush | Transistor | Zipper | Average |
|---|---|---|---|---|---|---|---|---|---|---|---|---|---|---|---|---|
| ST [19] | 98.80 | **99.00** | **99.30** | 97.40 | **97.20** | 98.80 | 95.50 | 98.30 | 98.50 | 97.60 | 97.80 | 98.30 | **98.90** | 82.50 | 98.50 | 97.09 |
| IKD [20] | 98.71 | 97.04 | 98.53 | 95.68 | 93.88 | **98.99** | **98.03** | 98.55 | 98.71 | 98.38 | 98.79 | 98.63 | 98.58 | **97.13** | 97.56 | 97.81 |
| SPADE [9] | 97.50 | 93.70 | 97.60 | 87.40 | 88.50 | 98.40 | 97.20 | **99.00** | 99.10 | 98.10 | 96.50 | 98.90 | 97.90 | 94.10 | 96.50 | 96.03 |
| PaDiM [8] | **99.10** | 97.30 | 99.20 | 94.10 | 94.90 | 98.30 | 96.70 | 98.50 | 98.20 | 97.20 | 95.70 | 98.50 | 98.80 | 97.50 | **98.50** | 97.50 |
| **CDO** | 99.08 | 98.40 | 99.17 | 97.20 | 95.85 | 99.30 | 97.60 | 98.64 | 99.24 | 98.54 | 98.94 | 99.01 | 98.86 | 95.30 | 98.21 | **98.22**± 0.05 |
| CFLOW* [6] | **99.25** | 98.99 | 99.66 | 98.01 | 96.65 | 98.98 | **97.64** | 98.98 | 98.89 | 98.56 | 98.95 | 98.86 | 98.93 | **97.99** | 99.08 | 98.63 |
| **CDO*** | 99.11 | **99.36** | **99.73** | **98.48** | **97.66** | **99.30** | 97.60 | 98.64 | **99.42** | **98.59** | **99.21** | **99.40** | **99.24** | 95.53 | **99.19** | **98.70**± 0.16 |

TABLE II. QUANTITATIVE COMPARISON OF ANOMALY DETECTION METHODS ON THE MVTEC2D DATASET IN TERMS OF AU-PRO. CDO* AND CFLOW* DENOTE THE EVALUATED RESULTS UNDER THE EVALUATION SETTING OF CFLOW.

| Category | Carpet | Grid | Leather | Tile | Wood | Bottle | Cable | Capsule | Hazelnut | Metal Nut | Pill | Screw | Toothbrush | Transistor | Zipper | Average |
|---|---|---|---|---|---|---|---|---|---|---|---|---|---|---|---|---|
| ST [19] | 95.80 | **96.60** | 98.00 | 92.10 | **93.60** | 95.10 | 87.70 | 92.20 | 94.30 | 94.50 | 96.50 | 93.00 | 92.20 | 69.50 | 95.20 | 92.42 |
| IKD [20] | 94.49 | 87.73 | 97.64 | 86.35 | 89.06 | 96.08 | **94.21** | 90.62 | 95.97 | 94.69 | 96.09 | 92.95 | 87.01 | **93.78** | 91.55 | 92.55 |
| SPADE [9] | 94.70 | 86.70 | 97.20 | 75.90 | 87.40 | 95.50 | 90.90 | **93.70** | 95.40 | 95.40 | 94.60 | **96.00** | 93.50 | 87.40 | 92.60 | 91.79 |
| PaDiM [8] | 96.20 | 94.60 | 97.80 | 86.00 | 91.10 | 94.80 | 88.80 | 93.50 | 92.60 | 85.60 | 92.70 | 94.40 | 93.10 | 84.50 | **95.90** | 92.11 |
| **CDO** | **96.77** | 96.02 | **98.34** | 90.51 | 92.87 | **97.17** | 94.17 | 92.97 | **97.39** | 95.74 | 96.59 | 94.33 | 90.50 | 92.56 | 94.28 | **94.68**± 0.20 |
| CFLOW* [6] | **97.70** | 96.08 | **99.35** | 94.34 | 95.79 | 96.80 | 93.53 | 93.40 | 96.68 | 91.65 | 95.39 | 95.30 | 95.06 | 81.40 | **96.60** | 94.60 |
| **CDO*** | 96.77 | **97.83** | 99.10 | **94.63** | **96.22** | **97.66** | **94.17** | **96.63** | **98.98** | **96.69** | **97.01** | **96.99** | **95.98** | **92.54** | 96.24 | **96.50**± 0.31 |

TABLE III. QUANTITATIVE COMPARISON OF ANOMALY DETECTION METHODS ON THE MVTEC3D DATASET IN TERMS OF AU-PRO.

| Modality | Category | Bagel | Cable gland | Carrot | Cookie | Dowel | Foam | Peach | Potato | Rope | Tire | Average |
|---|---|---|---|---|---|---|---|---|---|---|---|---|
| Voxel | GAN [1] | 44.00 | 45.30 | 82.50 | 75.50 | 78.20 | 37.80 | 39.20 | 63.90 | 77.50 | 38.90 | 58.28 |
| | AE [1] | 26.00 | 34.10 | 58.10 | 35.10 | 50.20 | 23.40 | 35.10 | 65.80 | 1.50 | 18.50 | 34.78 |
| | VM [1] | 45.30 | 34.30 | 52.10 | 69.70 | 68.00 | 28.40 | 34.90 | 63.40 | 61.60 | 34.60 | 49.23 |
| Depth | GAN [1] | 11.10 | 7.20 | 21.20 | 17.40 | 16.00 | 12.80 | 0.30 | 4.20 | 44.60 | 7.50 | 14.23 |
| | AE [1] | 14.70 | 6.90 | 29.30 | 21.70 | 20.70 | 18.10 | 16.40 | 6.60 | 54.50 | 14.20 | 20.31 |
| | VM [1] | 28.00 | 37.40 | 24.30 | 52.60 | 48.50 | 31.40 | 19.90 | 38.80 | 54.30 | 38.50 | 37.37 |
| PCD | 3D-ST_64 [25] | 93.90 | 44.00 | 98.40 | 90.40 | 87.60 | 63.30 | 93.70 | **98.90** | 96.70 | 50.70 | 81.76 |
| | 3D-ST_128 [25] | 95.00 | 48.30 | **98.60** | **92.10** | 90.50 | 63.20 | 94.50 | 98.80 | **97.60** | 52.40 | 83.10 |
| RGB | ST [19] | 93.22 | 90.95 | 97.48 | 91.73 | 89.80 | 69.82 | 93.30 | 95.49 | 94.54 | 88.41 | 90.47 |
| | **CDO** | **97.52** | **98.30** | 98.08 | 86.32 | **97.60** | **70.53** | **98.61** | 96.05 | 97.05 | **97.40** | **93.75**± 0.13 |

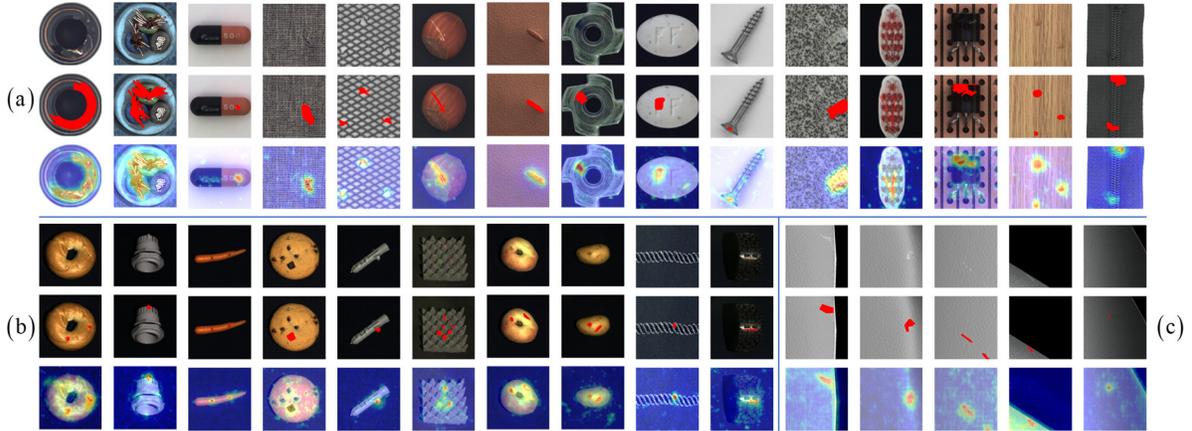

Fig. 4. Qualitative results of the proposed CDO on the MVTec2D (a), MVTec3D (b) and the collected dataset of plastic parts (c). Original images, ground truth masks and anomaly maps are listed from top to bottom.

### B. Comparisons with State-of-the-art methods

*MVTec2D:* Table I and Table II summarize per-class comparisons between CDO and other state-of-the-art models, namely, ST [20], IKD [21], SPADE [10], PaDiM [9], and CFLOW [7]. Although supervised methods can be directly applied using perturbed anomalies, they do not perform well as they are usually based on the closed-set assumption. Thus, they try to extract distinctive features for both normal and abnormal data and determine whether test data are more likely to be normal or abnormal. Therefore, they can only detect anomalies that occurred in the training set. However, the synthetic anomalies are visually different from real anomalies, and supervised methods perform weakly, so this study does not compare supervised segmentation methods. For the common setting, despite slight drops in some categories, CDO achieves 1.11% AU-ROC and 2.26% AU-PRO higher than its strong baseline ST, especially in the transistor category, with improvements of 12.8% AU-ROC and 11.14% AU-PRO. It may demonstrate that CDO better captures global context information than ST and effectively mitigates the overgeneralization problem. CDO also outperforms the existing best-performing method PaDiM by a large margin, particularly in AU-PRO (2.57%).

For comparison with CFLOW [7], we also report the performance with the same evaluation setting as CFLOW, which picks the best results of each category with various backbones and resolutions. With the CFLOW setting, CDO achieves 0.07% AU-ROC and 1.9% AU-PRO higher than CFLOW. Besides, the performance of CDO using CFLOW setting is further improved compared with using the common setting, indicating that the resolution and backbone should be cautiously selected in practical applications, as a suitable resolution and backbone performs better than a general one.

Moreover, in nine categories, CDO achieves nearly perfect AU-ROC (>99%). As shown in Table I and Table II, CDO achieves the highest performance in terms of AU-ROC and AU-PRO for both settings, demonstrating that CDO has the strongest ability for anomaly localization. Some qualitative results are shown in Fig. 4 (a), which shows that CDO can detect a variety of defects reliably.

*MVTec3D:* Table III summarizes per-class comparisons between CDO and other methods, including baselines [1] and 3D-ST [28]. We also evaluate ST [18] for a fair comparison. Among these methods, ST and CDO use RGB information, while others use depth information in voxel, depth, or point cloud for anomaly localization. CDO outperforms the best-performing depth information-based method 3D-ST by a large margin (10.72% AU-PRO), demonstrating that RGB information is also effective for geometric defect detection. The reason may be that geometric defects usually bring changes in RGB values, while surface texture defects sometimes do not result in depth changes, and depth-only methods may suffer from detecting these texture defects. CDO still outperforms its baseline ST in MVTec3D, illustrating the generality of CDO. Some qualitative results are shown in Fig. 4 (b). Whereas the depth and RGB information may complement each other well and produce better localization performance, a 3D sensor incurs additional costs. Therefore, it may be more appropriate to use a common RGB camera in some geometric defect localization tasks that do not require high precision.

### C. Ablation Studies

In this subsection, comprehensive experiments are conducted to show the influence of individual components of CDO. Different combinations of MOM and OOM are evaluated. Table IV shows the evaluated results on MVTec2D and MVTec3D of several instantiated cases. For cases without MOM, only discrepancies between normal FDs are optimized. Fig. 5 (a) shows the DDs of various combinations in the leather category of MVTec2D.

TABLE IV. QUANTITATIVE COMPARISON OF SEVERAL INSTANTIATED CASES. IN EVERY DATASET, THE LEFT COLUMN REPORTS AVERAGE AU-ROC, AND THE RIGHT COLUMN REPORTS AVERAGE AU-PRO. M DENOTES THE MOM. O DENOTES THE OOM. BEST IN BOLD.

| N | M | O | MVTec2D | | MVTec3D | |
|---|---|---|---|---|---|---|
| ① |   |   | 96.94± 0.12 | 91.37± 0.29 | 97.72± 0.05 | 93.11± 0.19 |
| ② |   | ✓ | 97.88± 0.03 | 93.71± 0.16 | **97.88± 0.05** | 93.52± 0.21 |
| ③ | ✓ |   | 97.94± 0.08 | 93.66± 0.28 | 97.78± 0.03 | 93.18± 0.18 |
| ④ | ✓ | ✓ | **98.25± 0.05** | **94.75± 0.20** | **97.88± 0.05** | **93.75± 0.13** |

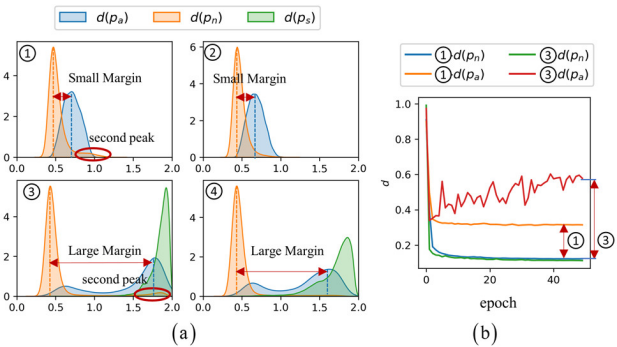

Fig. 5. (a) Discrepancy distributions of the corresponding cases. X-axis: $d(p)$, Y-axis: Density (b) The curves of the discrepancies over epoch in cases 1 and 3.

*Influence of MOM:* The influence of MOM can be concluded through the comparisons between Cases 1 and 3, and between Cases 2 and 4. Both comparisons show an improvement with MOM, especially in MVTec2D, with 2.33% and 1.04% AU-PRO gain, respectively.

From comparisons between Cases 1 and 3, and between Cases 2 and 4 as shown in Fig. 5 (a), the influence of the MOM can be deduced. Concretely, the margins in Cases 3 and 4 are significantly larger than those in Cases 1 and 2, as discrepancies between real abnormal FDs are implicitly maximized with the assistance of synthetic abnormal FDs in Cases 3 and 4. Fig. 5 (b) shows the optimizing processes of normal and abnormal average discrepancies of Cases 1 and 3 during training. Both average discrepancies between normal FDs and abnormal FDs in Case 1 fall off quickly and then settle at low values, and the final margins are small. In Case 3 with MOM, the average discrepancies between abnormal FDs drop first and then increase to high values, resulting in a much larger margin than that in Case 1. The curves in Fig. 5 (b) effectively prove that the overgeneralization problem exists in previous methods, while MOM can alleviate it and secure reliable predictions.

*Influence of OOM:* The influence of OOM can be concluded through the comparisons between Cases 1 and 2, and between Cases 3 and 4. As shown in Table IV, significant improvements can be obtained with OOM.

From the comparisons between Cases 1 and 2, and between Cases 3 and 4 as shown in Fig. 5 (a), the influence of OOM can be deduced. For the cases without OOM, small second peaks of discrepancies exist, as shown in Fig. 5 (a), which indicates that some tailed normal features are not aligned well between the expert and apprentice domains. OOM raises abundant attention on the tailed samples and removes those second peaks, decreasing the overlap between normal and abnormal DDs.

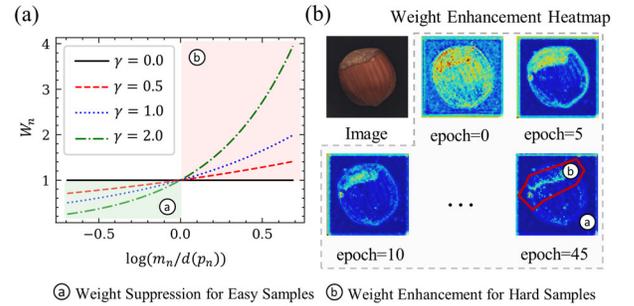

ⓐ Weight Suppression for Easy Samples  ⓑ Weight Enhancement for Hard Samples

Fig. 6. (a) Influence of $\gamma$ on the modulating factor of normal samples. (b) The heatmaps of weights under different epochs.

According to the comparison between discrepancies and the average, samples can be categorized into easy and hard samples. Fig. 6 (a) shows that a larger $\gamma$ in Eq. (8) will produce more significant suppression to easy samples and enhancement to hard samples. Fig. 6 (b) shows the heatmaps of weights calculated by Eq. (8) in the training procedure. OOM gradually assigns some pixels with significantly larger weights, *i.e.*, weights of tailed samples are enhanced.

To evaluate the influence of $\gamma$, this study conducts several experiments with different $\gamma$. Fig. 7 reports the evaluation results regarding the average of corresponding categories.

When $\gamma = 0$, OOM makes no effect, and the anomaly localization performance is subpar. The localization performance is improved first and then stabilizes as $\gamma$ is continuously increased. A suitable $\gamma$ can improve nearly 0.5% AU-ROC and 1.2% AU-PRO than $\gamma = 0$. More significant improvements are achieved in the object categories than in the texture categories, as normal patterns are more complex in the object categories, and OOM helps to optimize those tailed hard samples better.

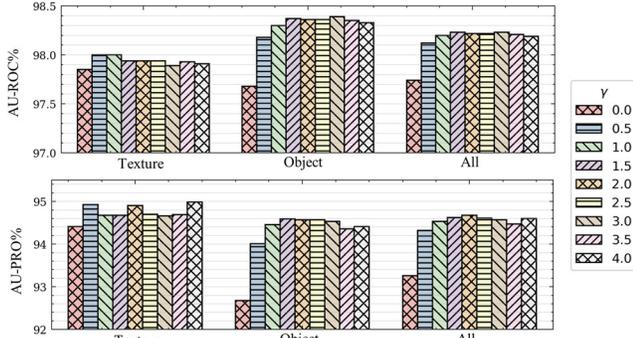

Fig. 7. The AU-ROC and AU-PRO of the proposed method in MVTec2D under different $\gamma$.

*Comparison of different backbones and resolutions:* Backbones and image resolutions significantly influence localization performance [7], [8]. This study compares different backbones, including HR18 (HRNet18), HR32, HR48 [27], Res18 (ResNet18), Res34, Res50 [29], and WRes50 [30] under two resolutions, $256 \times 256$ and $512 \times 512$. As Fig. 8 shows, ResNet under the resolution of $256 \times 256$ performs better in texture categories. By contrast, HRNet under the resolution of $256 \times 256$ performs better in object categories. The reason may be that anomaly localization on texture categories needs low-level structural information while high-level semantic information is needed on object categories, and HRNet can extract more semantic features owing to its efficient information transmission between its several branches. HR32 with a resolution of $512 \times 512$ has the highest average AU-PRO of 95.21%, and HN48 with a resolution of $256 \times 256$ has the highest average AU-ROC of 98.28%.

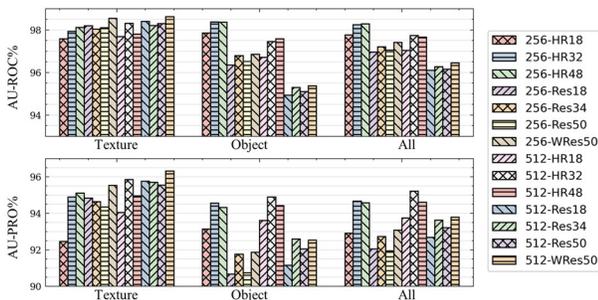

Fig. 8. The AU-ROC and AU-PRO of the proposed method in MVTec2D with different backbones and resolutions.

### D. Complexity Evaluation

This study evaluates the actual computation and memory complexity for the proposed CDO and other state-of-the-art methods, including SPADE [10], PaDiM [9], and CFLOW [7] in terms of inference speed and model size metrics.

TABLE V. COMPLEXITY COMPARISON IN TERMS OF INFERENCE SPEED (FPS) AND MODEL SIZE (MB).

| Method | Backbone | Inference speed | Model size |
|---|---|---|---|
| **CDO** | Res18 | 18.70 | 89 |
| | HR32 | 16.58 | 227 |
| | WRes50 | 18.35 | 526 |
| PaDiM [8] | WRes50 | 0.47 | 4624 |
| SPADE [9] | WRes50 | 0.39 | 1284 |
| CFLOW [6] | Res18 | 12.72 | 96 |
| | WRes50 | 10.02 | 902 |

The model size is defined as the size of all floating-point parameters in the corresponding model in Table V. SPADE needs to store a feature gallery, and PaDiM needs to store pixel-wise covariance, so they require high memory consumption, and both CFLOW and CDO require less memory than SPADE and PaDiM. In addition, since the decoder in CFLOW is larger than the apprentice feature extractor $\phi_A$ in CDO, CDO has a smaller model size than CFLOW. Ultimately, CDO requires the smallest size and is 1.1× to 1.7× smaller than CFLOW [7].

The inference speed shown in Table V is measured using a computer with Intel i7@2.30GHz CPU, NVIDIA GTX 2060 GPU, and 16G RAM. All models are allocated to GPU to accelerate the inference. The data input time is taken into consideration for the evaluation of each method when measuring the inference speed. CDO can run in real-time and is 41.9× faster than SPADE and 34.7× faster than PaDiM. Besides, as the extractors $\phi_E$ and $\phi_A$ of CDO can work in parallel, while the encoder and the decoder work sequentially in CFLOW, CDO is 1.3× to 1.7× faster than CFLOW. Notably, the inference speed of CDO is almost unchanged under different backbones as the main bottleneck of the inference speed is the data input time. Excluding the data input time, our model can achieve a very high computational speed, around 80 fps with Res18. Overall, CDO with HR32 has the best localization performance and considerable performance on inference speed and memory consumption.

### E. Application to Automotive Plastic Parts Inspection

To further evaluate the practical performance of CDO, we apply CDO to the inspection of real-world automotive plastic parts. The inspection device is shown in Fig. 9 (a). An automotive plastic parts dataset was collected using the device, and the dataset contains 1500 normal patches for training, 500 normal patches, and 271 abnormal patches for testing. All samples have a resolution of $256 \times 256$. Compared to existing anomaly localization datasets like MVTec2D [3], this dataset is much more challenging because the normal patches have large inter-class variance, and the anomalies are extremely small in the dataset. Some samples are shown in Fig. 9 (b). This study compares the performance of our proposed CDO with different backbones and its strong baseline ST [20].

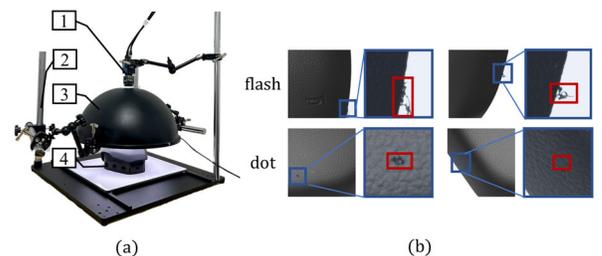

Fig. 9. (a) The device for inspection of automotive plastic parts. 1: camera, 2: stand, 3: light source, 4: the inspected plastic part. (b) Some samples from the collected dataset.

TABLE VI. QUANTITATIVE COMPARISON OF ST AND THE PROPOSED CDO WITH DIFFERENT BACKBONES. BEST IN BOLD.

| Method | Backbone | AU-PRO |
|---|---|---|
| ST [19] | Res18 | 82.60±1.30 |
| CDO | Res18 | 84.55±1.31 |
|  | Res34 | 87.15±0.50 |
|  | Res50 | 86.30±1.52 |
|  | **WRes50** | **90.30±0.32** |
|  | HR18 | 85.14±0.73 |
|  | HR32 | 87.86±0.54 |
|  | HR48 | 87.94±0.78 |

As mentioned earlier, WRes50 performs better for texture images, and the dataset collected is more like texture images. Hence, the proposed CDO performs the best with WRes50 and has 7.7% higher AU-PRO than the baseline ST [20], as shown in Table VI. Some examples of the localization results can be found in Fig. 4 (c). Despite the hard-to-detect anomalies, CDO localizes them impressively.

In summary, all experiment results show that the proposed CDO achieves a state-of-the-art performance and performs excellently on the automotive plastic parts dataset.

## V. CONCLUSION

Most existing unsupervised anomaly localization methods focus on optimizing the discrepancies between normal feature distributions (FDs) in the expert and apprentice domains and neglect the optimization of abnormal FDs, and the high generalization abilities of convolutional neural network may result in overgeneralization. This paper proposes a collaborative discrepancy optimization (CDO) method to mitigate the overgeneralization problem in two aspects. First, a margin optimization module is proposed to explicitly enlarge the margin between the discrepancy distributions (DDs) with the assistance of synthetic anomalies. Second, an overlap optimization module is developed to reduce the overlap between DDs by assigning weights to individual samples dynamically. The experiments on MVTec2D and MVTec3D demonstrate that the proposed CDO achieves a state-of-the-art anomaly localization performance with very fast computation speed and low memory consumption. The application in real-world automotive plastic parts inspection further proves its effectiveness.

In the future, better anomaly generation methods like generative adversarial networks can be investigated to replace random perturbation in order to generate more real synthetic abnormal FDs and provide better assistance to optimize real abnormal DDs. On the other hand, the proposed image anomaly localization methods can be deployed to other industrial applications, such as intelligent fault diagnosis with thermal images.

## REFERENCES

[1] P. Bergmann and D. Sattlegger, "The MVTec 3D-AD Dataset for Unsupervised 3D Anomaly Detection and Localization," in *International Joint Conference on Computer Vision, Imaging and Computer Graphics Theory and Applications,* 2022.

[2] H. Dong, K. Song, Y. He, J. Xu, Y. Yan, and Q. Meng, "PGA-Net: Pyramid Feature Fusion and Global Context Attention Network for Automated Surface Defect Detection," *IEEE Trans. Ind. Informatics*, vol. 16, no. 12, pp. 7448–7458, 2020, doi: 10.1109/TII.2019.2958826.

[3] P. Bergmann, K. Batzner, M. Fauser, D. Sattlegger, and C. Steger, "The MVTec Anomaly Detection Dataset: A Comprehensive Real-World Dataset for Unsupervised Anomaly Detection," *Int. J. Comput. Vis.*, vol. 129, no. 4, pp. 1038–1059, 2021, doi: 10.1007/s11263-020-01400-4.

[4] G. Liu, W. Shen, L. Gao, and A. Kusiak, "Knowledge transfer in fault diagnosis of rotary machines," *IET Collab. Intell. Manuf.*, vol. 4, no. 1, pp. 17–34, 2022, doi: 10.1049/cim2.12047.

[5] A. Glowacz, "Fault diagnosis of electric impact drills using thermal imaging," *Meas. J. Int. Meas. Confed.*, vol. 171, no. August 2020, p. 108815, 2021, doi: 10.1016/j.measurement.2020.108815.

[6] M. Rudolph, B. Wandt, and B. Rosenhahn, "Same Same But DifferNet: Semi-Supervised Defect Detection with Normalizing Flows," in *Proceedings of the IEEE Winter Conference on Applications of Computer Vision*, 2021, pp. 1906–1915, doi: 10.1109/wacv48630.2021.00195.

[7] D. Gudovskiy, S. Ishizaka, and K. Kozuka, "CFLOW-AD: Real-Time Unsupervised Anomaly Detection with Localization via Conditional Normalizing Flows," in *Proceedings of the IEEE Winter Conference on Applications of Computer Vision*, 2021, pp. 98–107.

[8] Q. Wan, L. Gao, X. Li, and L. Wen, "Industrial Image Anomaly Localization Based on Gaussian Clustering of Pre-trained Feature," *IEEE Trans. Ind. Electron.*, vol. 0046, 2021, doi: 10.1109/tie.2021.3094452.

[9] T. Defard, A. Setkov, A. Loesch, and R. Audigier, "PaDiM: A Patch Distribution Modeling Framework for Anomaly Detection and Localization," in *International Conference on Pattern Recognition*, 2021, pp. 475–489, doi: 10.1007/978-3-030-68799-1_35.

[10] T. Reiss, N. Cohen, L. Bergman, and Y. Hoshen, "PANDA: Adapting Pretrained Features for Anomaly Detection and Segmentation," in *Proceedings of the IEEE Conference on Computer Vision and Pattern Recognition*, 2021, pp. 2805–2813, doi: 10.1109/cvpr46437.2021.00283.

[11] J. Yi and S. Yoon, "Patch SVDD: Patch-Level SVDD for Anomaly Detection and Segmentation," in *Proceedings of the Asian Conference on Computer Vision*, 2020, pp. 375–390, doi: 10.1007/978-3-030-69544-6_23.

[12] V. Zavrtanik, M. Kristan, and D. Skočaj, "Reconstruction by inpainting for visual anomaly detection," *Pattern Recognit.*, vol. 112, 2021, doi: 10.1016/j.patcog.2020.107706.

[13] Y. Shi, J. Yang, and Z. Qi, "DFR: Deep Feature Reconstruction for Unsupervised Anomaly Segmentation," *Neurocomputing*, vol. 424, pp. 9–22, 2021, doi: 10.1016/j.neucom.2020.11.018.

[14] V. Zavrtanik, M. Kristan, and D. Skočaj, "DRAEM -- A discriminatively trained reconstruction embedding for surface anomaly detection," in *Proceedings of the IEEE International Conference on Computer Vision*, 2021, pp. 8310–8319, doi: 10.1109/ICCV48922.2021.00822.

[15] X. Tao, D. P. Zhang, W. Ma, Z. Hou, Z. Lu, and C. Adak, "Unsupervised Anomaly Detection for Surface Defects with Dual-Siamese Network," *IEEE Trans. Ind. Informatics*, vol. 3203, no. c, 2022, doi: 10.1109/TII.2022.3142326.

[16] D. Gong et al., "Memorizing normality to detect anomaly: Memory-augmented deep autoencoder for unsupervised anomaly detection," in *Proceedings of the IEEE International Conference on Computer Vision*, 2019, pp. 1705–1714, doi: 10.1109/ICCV.2019.00179.

[17] P. Bergmann, S. Löwe, M. Fauser, D. Sattlegger, and C. Steger, "Improving unsupervised defect segmentation by applying structural similarity to autoencoders," in *International Joint Conference on Computer Vision, Imaging and Computer Graphics Theory and Applications*, 2019, vol. 5, pp. 372–380, doi: 10.5220/0007364503720380.

[18] M. Salehi, N. Sadjadi, S. Baselizadeh, M. H. Rohban, and H. R. Rabiee, "Multiresolution Knowledge Distillation for Anomaly Detection," in *Proceedings of the IEEE Conference on Computer Vision and Pattern Recognition*, 2021, pp. 14897–14907, doi: 10.1109/CVPR46437.2021.01466.

[19] P. Bergmann, M. Fauser, D. Sattlegger, and C. Steger, "Uninformed Students: Student-Teacher Anomaly Detection with Discriminative

Latent Embeddings," *in Proceedings of the IEEE Conference on Computer Vision and Pattern Recognition*, 2020, pp. 4182–4191, doi: 10.1109/CVPR42600.2020.00424.

[20] G. Wang, S. Han, E. Ding, and D. Huang, "Student-Teacher Feature Pyramid Matching for Anomaly Detection," *in Proceedings of the British Machine Vision Conference*, 2021

[21] Y. Cao, Q. Wan, W. Shen, and L. Gao, "Informative knowledge distillation for image anomaly segmentation," *Knowledge-Based Syst.*, 2022, doi: 10.1016/j.knosys.2022.108846.

[22] H. Deng and X. Li, "Anomaly Detection via Reverse Distillation from One-Class Embedding," in IEEE Conference on Computer Vision and Pattern Recognition, 2022, pp. 9727--9736, doi: 10.48550/arXiv.2201.10703.

[23] Y. Cao, Y. Song, X. Xu, S. Li, Y. Yu, and W. Shen, "Semi-supervised Knowledge Distillation for Tiny Defect Detection," in *International Conference on Computer Supported Cooperative Work in Design*, 2022, pp. 1010–1015, doi: 10.1109/CSCWD54268.2022.9776026.

[24] T. Y. Lin, P. Goyal, R. Girshick, K. He, and P. Dollar, "Focal Loss for Dense Object Detection," *IEEE Trans. Pattern Anal. Mach. Intell.*, vol. 42, no. 2, pp. 318–327, 2020, doi: 10.1109/TPAMI.2018.2858826.

[25] Jia Deng, Wei Dong, R. Socher, Li-Jia Li, Kai Li, and Li Fei-Fei, "ImageNet: A large-scale hierarchical image database," *in Proceedings of the IEEE Conference on Computer Vision and Pattern Recognition*, 2009, pp. 248–255, doi: 10.1109/cvprw.2009.5206848.

[26] Loshchilov and F. Hutter, "Decoupled Weight Decay Regularization," *in Proceedings of the International Conference on Learning Representations*, 2019.

[27] J. Wang et al., "Deep High-Resolution Representation Learning for Visual Recognition," *IEEE Trans. Pattern Anal. Mach. Intell.*, vol. 43, no. 10, pp. 3349–3364, 2020, doi: 10.1109/tpami.2020.2983686.

[28] P. Bergmann and D. Sattlegger, "Anomaly Detection in 3D Point Clouds using Deep Geometric Descriptors," *ArXiv Prepr.*, pp. 1–21.

[29] K. He, X. Zhang, S. Ren, and J. Sun, "Deep residual learning for image recognition," *in Proceedings of the IEEE Conference on Computer Vision and Pattern Recognition*, 2016, pp. 770–778, doi: 10.1109/CVPR.2016.90.

[30] B. S. Zagoruyko and N. Komodakis, "Wide Residual Networks," *in Proceedings of the British Machine Vision Conference*, 2016.

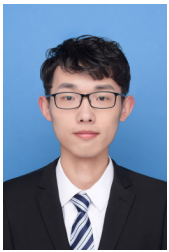

**Yunkang Cao** (Student Member, IEEE) was born in Jiangxi, China, in 1999. He received the B.S. degree from Huazhong University of Science and Technology (HUST), Wuhan, China, in 2020, where he is currently pursuing the Ph.D. degree in mechanical engineering.

His current research intesests include deep learning, anomaly detection and related applications in real industrial scenarios.

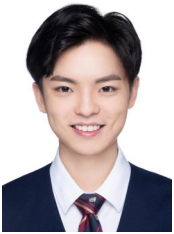

**Xiaohao Xu** received his B.S. degree in mechanical design, manufacturing and automation from Huazhong University of Science and Technology (HUST), Wuhan, China in 2022.

His current research interests include the fundamental theory and real-world applications of robotics, computer vision, and video understanding.

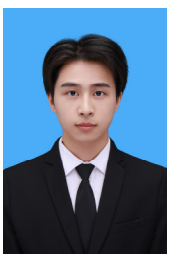

**Zhaoge Liu** was born in Henan, China, in 1999. He received his B.S. degree in mechanical design, manufacturing and automation from Central South University (CSU) in Changsha, China in 2021 and is currently pursuing the M.S. degree in mechanical engineering from Huazhong University of Science and Technology (HUST) in Wuhan, China.

His current research interests include deep learning, anomaly detection, knowledge distillation.

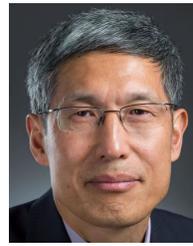

**Weiming Shen** (Fellow, IEEE) received the B.E. and M.S. degrees in mechanical engineering from Northern Jiaotong University, Beijing, China, in 1983 and 1986, respectively, and the Ph.D. degree in system control from the University of Technology of Compiegne, Compiegne, France, in 1996. He is currently a Professor with the Huazhong University of Science and Technology (HUST), Wuhan, China, and an Adjunct Professor with the University of Western Ontario, London, ON, Canada. Before joining HUST in 2019, he was a Principal Research Officer at the National Research Council Canada. He is a Fellow of Canadian Academy of Engineering and the Engineering Institute of Canada.

His work has been cited more than 16 000 times with an h-index of 61. He authored or coauthored several books and more than 560 articles in scientific journals and international conferences in related areas. His research interests include agent-based collaboration technologies and applications, collaborative intelligent manufacturing, the Internet of Things, and Big Data analytics